Systems Article

# Mission-level Robustness with Rapidly-deployed, Autonomous Aerial Vehicles by Carnegie Mellon Team Tartan at MBZIRC 2020

**Anish Bhattacharya**, **Akshit Gandhi**, **Lukas Merkle**, **Rohan Tiwari**,
**Karun Warrior**, **Stanley Winata**, **Andrew Saba**, **Kevin Zhang**, **Oliver Kroemer**
and **Sebastian Scherer**

Robotics Institute, Carnegie Mellon University, Pittsburgh, PA 15213

**Abstract:** For robotic systems to succeed in high risk, real-world situations, they have to be quickly deployable and robust to environmental changes, under-performing hardware, and mission subtask failures. These robots are often designed to consider a single sequence of mission events, with complex algorithms lowering individual subtask failure rates under some critical constraints. Our approach utilizes common techniques in vision and control, and encodes robustness into mission structure through outcome monitoring and recovery strategies. In addition, our system infrastructure enables rapid deployment and requires no central communication. This report also includes lessons in rapid field robotic development and testing. We developed and evaluated our systems through real-robot experiments at an outdoor test site in Pittsburgh, Pennsylvania, USA, as well as in the 2020 Mohamed Bin Zayed International Robotics Challenge. All competition trials were completed in fully autonomous mode without RTK-GPS. Our system placed fourth in Challenge 2 and seventh in the Grand Challenge, with notable achievements such as popping five balloons (Challenge 1), successfully picking and placing a block (Challenge 2), and dispensing the most water onto an outdoor, real fire with an autonomous UAV (Challenge 3).

## 1. Introduction

Autonomous field robots have the potential to reduce risk, improve efficiency, increase precision, and transform the way tasks are performed in diverse sectors, including security, construction, and emergency response. Realizing this vision will require robots to function in outdoor, unknown environments and be robust to changing conditions, faulty hardware, and failed mission subtasks. While these issues are typically not considered in algorithm development, the MBZIRC 2020 (Mohamed Bin Zayed International Robotics Challenge, 2020) emphasizes them explicitly. The competition prompts teams to develop fully autonomous systems that can, without a priori maps of the arenas and only minimal onsite tuning time, sense their surroundings and precisely manipulate objects. All system computation had to be self contained aboard the robots, with minimal communication, since reliable network service could not be assured in the large, outdoor arenas.

MBZIRC 2020 featured three challenges, with the unmanned aerial vehicle (UAV) tasks as follows. Challenge 1 focused on airspace safety with two tasks: popping helium-filled balloons tied to vertical poles, and capturing a foam ball suspended by a UAV flying in a figure-8 trajectory. Challenge 2 focused on construction, where UAVs build an elevated wall structure with foam blocks. Challenge 3 focused on firefighting, with UAVs tackling indoor and outdoor fires. The final, Grand Challenge included tasks from all three challenges to be completed simultaneously. CMU Team Tartan's development of aerial robot systems for completing these tasks is described in this paper.

Interest in fully-autonomous, small, unmanned aerial systems (sUAS) has rapidly increased in recent years with the availability of affordable UAV platforms and the greater capabilities of small, onboard computers. However, much of sUAS field research has focused on remote sensing, rather than the aerial manipulation tasks that MBZIRC calls for. This is largely due to shared limitations of all sUAS: UAVs are not inherently stable platforms, so any contact with foreign objects or the environment must be very precise and deliberate. Furthermore, the limited payload capacity prevents the use of modern, feature-rich robotic manipulators. State-of-the-art field sUAS are also often very task-specific, relying on expensive sensors and complex algorithms to maximize the success rate in their one particular task. While this approach is useful under known and unchanging conditions, MBZIRC requires quick system adaptation, easy problem diagnosing under tight time constraints, and consistent performance, regardless of hardware faults or changing environmental conditions.

The key ideas in this work emphasize lightweight and robust autonomy for aerial manipulation tasks. Specifically, we develop UAV-friendly manipulators that require no additional motors or mechanized parts and, as such, are fit for rough, repeated use on a sUAS. We also design our system around commonly-used and low-cost sensors, such as cameras and laser rangefinders, and simple subtask algorithms that lend themselves to repeated use, thus enhancing mission-level robustness.

Our main contributions include our development strategies of simple, yet robust algorithms in computer vision, vision-based control and guidance, and mission planning. We use GPS and onboard cameras for state estimation but no external positioning stations such as real-time kinematic GPS (RTK-GPS). We present an autonomous system-startup and mission-launching architecture that facilitates the rapid deployment of our robots in time-constrained and WiFi-denied settings. We show that mission plans containing self-check cycles helped us overcome faulty hardware and failed subtasks and still succeed in the overall mission. We describe the modularity in our software and hardware approach and some critical lessons learned during the field development and testing cycle. We also thoroughly assess failure cases observed in the MBZIRC trials and point out some immediate potential improvements. We hope that this report will be useful to future teams competing in robotics competitions and also for researchers building robotics systems for use in outdoor, semi-structured environments.

The structure of this paper is as follows. Section 2 reviews related work on methods relevant for the MBZIRC 2020 tasks. Section 3 describes the common hardware and software systems used between the challenges. Core concepts in our work are described in Section 4. Each core concept and the following hardware, algorithms, and approaches are addressed in detail for Challenges 1, 2, and 3 in Sections 5, 6, and 7, respectively. Competition results are discussed in Section 8, and a summary of lessons learned from our work can be found in Section 9. Finally, Section 10 describes our conclusions. A video of our competition performance, as well as supplementary sections involving additional work and post-competition analysis can be found on our website[1].

---

[1] https://www.sites.google.com/view/cmuteamtartan

## 2. Related Work

In search for a target across large spaces, the University of Bonn's MBZIRC 2020 Challenge 1 (Beul et al., 2020) balloon search strategy composed of a coarse arena-length (100m) lawnmower path repeated multiple times if necessary (similarly, (Garcia et al., 2020), (Castaño et al., 2019), (Bähnemann et al., 2019), and (Spurný et al., 2019) use this scheme but apply various Dubins path and multi-robot optimizations). However, they calculate and store all visible balloons' 3-D locations, calculated from the relationship between pixel size and their known physical size. Note that getting a bird's-eye-view of the large arena and targets is not straightforward due to other higher-altitude UAVs operating simultaneously. Due to possible inaccuracies in long-range target position calculation, our method for target search instead uses a finer-resolution lawnmower path that must travel close to targets, but can be manually adjusted to focus on any particular portion of the arena.

Object detection onboard UAVs is generally a tradeoff between generalization and computational efficiency, or more specifically, machine learning (e.g., deep learning) versus classical methods (e.g., Hough transforms, edge detectors, color segmentation). Deep learning inference onboard these platforms is restricted to smaller architectures or optimized methods, such as single-shot detection using MobileNet (Howard et al., 2017) or TinyYOLOv3 (Redmon and Farhadi, 2018). (Beul et al., 2020) is interested in long-range balloon localization accuracy and used tensor processing unit (TPU) accelerated inference paired with a custom circle fitter to estimate the target contour, whereas (Garcia et al., 2020) used hue, saturation, value (HSV) filtering and detected contours in a depth image. For colored plate detection in MBZIRC 2017 Challenge 3, (Bähnemann et al., 2019) used a hand-tunable blob detector with a feature classifier which they tuned onsite, (Castaño et al., 2019) used HSV-space segmentation and clustering, and (Spurný et al., 2019) used an ellipse detection algorithm with an auto-calibrated Gaussian mixture model (GMM) for color segmentation. (Keipour et al., 2021b) also developed an ellipse detection algorithm for landing on a moving platform in MBZIRC 2017 Challenge 1. In contrast, our methods for colored balloon, ball, and block detection rely on deep learning-based inference to ensure good generalization to the previously-unseen arena scene with minimal tuning.

The autonomous construction task inspires various approaches. (Krizmancic et al., 2020) developed a decentralized task planning and coordination framework for a cooperative aerial-ground robot team for the wall building problem, demonstrated in simulation. The winners of MBZIRC 2020 Challenge 2 outline their approach in (Baca et al., 2020). Baca, et al., employed one UAV to create a topological arena map which was shared to the other UAVs. Block detection was done with HSV thresholding, and pickup was done with visual servoing control; block pickup success was monitored by checking the vehicle's estimated mass and attitude post-pickup. Wall detection for placement was done by generating wall candidates from a depth thresholding operation (using a RGB-D camera) then finding contours in the RGB image. While (Bähnemann et al., 2019) used position-based visual servoing (PBVS) methods for picking up colored plates in MBZIRC 2017, we use image-based visual servoing (IBVS) for block pickup and placement which is less sensitive to image noise and makes fewer 3-D assumptions. (Keipour et al., 2021a) also uses IBVS for landing on a moving platform in MBZIRC 2017. Stable aerial manipulation tasks might also be performed with fully-actuated UAVs (Rajappa et al., 2015; Keipour et al., 2020; Odelga et al., 2016).

Autonomous firefighting contains elements of infrastructure inspection, thermal and flame detection, and extinguishing. (Eschmann et al., 2012) examines the use of UAVs for generating high-resolution 2D imagery of buildings. (Krawczyk et al., 2015) uses thermal imagery on a UAV to detect a building's heat loss. Specific fire detection approaches are covered in (Yuan et al., 2015) and (Yuan et al., 2017), ranging from thermal histogram-based segmentation and the fire's optical flow calculation to color segmentation using the Lab color model's "a" channel. In (Han and Lin, 2016), a static background is subtracted such that a flame candidate remains, which is then classified based on color and shape. There remains limited work on the full surveil-detect-extinguish firefighting task with autonomous UAVs, which is what we tackle in this work.

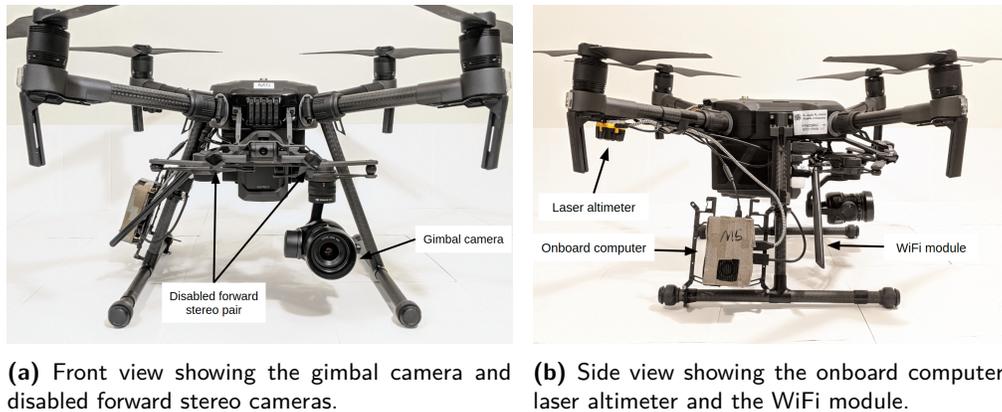

**(a)** Front view showing the gimbal camera and disabled forward stereo cameras.

**(b)** Side view showing the onboard computer, laser altimeter and the WiFi module.

**Figure 1.** Common UAV platform.

## 3. Common Platform

A common hardware and software platform was used across all three challenges, with some challenge-specific modifications.

### 3.1. Hardware design

Our common aerial robotic platform consisted of the DJI M210 V2 (DJI M210 V2 Specifications, 2020) without RTK-GPS, with challenge-specific modifications as described in later sections. A DJI Manifold 2-C (DJI Manifold 2-C Specifications, 2020) served as the onboard computer, mounted on one of the UAV's legs and wrapped in electromagnetic shielding cloth to minimize GPS interference. USB power banks provided reliable power to the additional USB cameras used in Challenges 2 and 3. Other notable USB devices common across all robots include a WiFi USB module (for minimal parameter tuning and monitoring), Terabee EVO 60m sensor (for accurate altitude estimation) (TeraRanger Evo 60m Specifications, 2020), and Blink(1) LED (for rapid deployment status indication) (Blink(1) USB, 2020). A DJI Zenmuse X5S gimbal camera (DJI Zenmuse X5S Specifications, 2020) is mounted to the front of the UAV. The built-in forward-facing stereo pair, typically used for automatic obstacle avoidance, was manually disabled by covering the lens with electrical tape. The available software switch did not work due to a bug that has since been acknowledged and fixed in future versions of the DJI SDK.

### 3.2. Software design

The common software system (Figure 2) uses Robot Operating System (ROS) and is modular; the state machine and task-specific sensor modules were unique to each challenge. The DJI Software Development Kit (SDK) (DJI Developer SDKs, 2020) state estimate uses the ultrasonic altimeter, inertial measurement unit (IMU), compass, GPS, and downward stereo cameras but was edited directly with our laser altimeter for a more precise altitude. We downsample the gimbal camera feed in the DJI SDK ROS code to reduce message lag and speed up downstream perception algorithm computation. The software system outputs velocity commands to the DJI SDK velocity controller either directly from the state machine (during local interaction) or through a cascaded trajectory controller (during global search).

### 3.3. Rapid deployment

We developed an automated rapid deployment pipeline that quickly, easily, and reliably starts and checks all necessary onboard systems and sensors before a field test or competition trial. This pipeline does not rely on any network communications and is triggered upon powering on, with the sequence managed by a watchdog node and the status externally visible by a Blink(1) USB LED. Full details and a diagram are available in Appendix A.

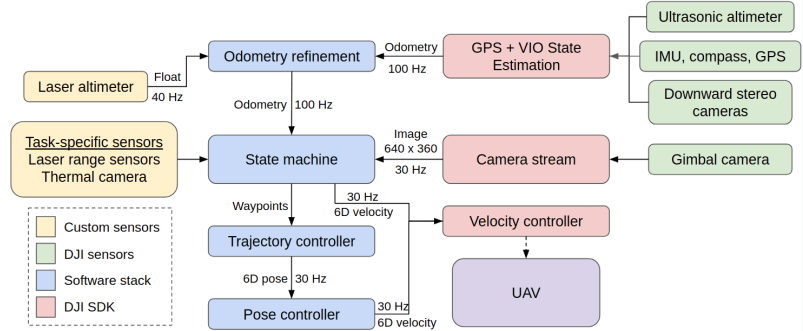

**Figure 2.** Common software system.

## 4. Core Concepts

Each system in this report was developed with several common, core concepts that contributed to the overall success and robustness observed in the development and testing process as well as the competition performance. This section outlines the theme of each core concept, while the development and realization of them for each challenge is explained in Sections 5, 6, and 7.

### 4.1. Hardware

As outlined in Section 3, a common UAV platform in hardware, software, and startup was used for every robotic system. This uniformity ensured that general capabilities such as takeoff/land, GPS waypoint following, and handling sensor data were available for all challenges. It also allowed hardware modularity, where parts may be swapped between robots and full robots can be quickly retrofitted for use in a different challenge, as needed. Two core concepts dictated the use of task-specific hardware: task-specific sensors and lightweight attachments.

#### 4.1.1. Task-specific sensors
Task-specific sensors were added to meet the particular needs of each challenge, including laser range sensors, a fisheye camera, and a thermal camera. Sensors provided additional feedback for monitoring the task and triggering recovery strategies. The same sensor types used for different tasks are connected to the larger system architecture using the same drivers and ROS nodes.

#### 4.1.2. Lightweight attachments
We used simple, lightweight, and task-specific payloads, such as the propellers themselves (popping balloons), a carbon fiber-mounted net (catching the ball), a passive electro-permanent gripper (picking and placing blocks), and a windshield-washer water pump (dispensing water).

### 4.2. Control strategies and architecture

Specific themes were pursued in mission structure and task monitoring for each challenge, which added significant, observed robustness to every mission during the competition.

#### 4.2.1. Global search and local interaction
As every task involves a mobile search followed by object interaction, we structured each mission with two distinct components. We first perform a coarse global search over pre-set areas of interest, then resume finer interactions with discovered local targets.

### 4.2.2. Outcome monitoring and recovery strategies

During local interaction with a target, the system monitors task progress and implements a recovery strategy if failure is detected. The recovery strategy, though simply a reset and re-attempt at the task, led to robust mission success through repeated task attempts.

## 4.3. Vision pipeline

Various cameras are used to detect objects and their features, as well as to provide real-time feedback for close interactions. The vision pipeline must be robust to changes in scenery, since we are not given prior visual data from the arenas. As described in Section 7.3, only thermal-target detection did not require such generalization, so in that case we chose to use simple pixel thresholding.

### 4.3.1. Robust detection in changing environments

The first step in our vision pipelines is to accurately detect target objects including a small and fast-moving ball, swaying balloons, and different-colored blocks. Our robust detection methods involved trained object detection networks to ensure that target localization is invariant to lighting, weather, and surroundings.

### 4.3.2. Continuous visual feedback for precise interactions

After object detection, consistent visual feedback is used by the onboard controllers during local interaction with a target. This is aided by the use of a gimbal camera, and must be able to withstand external disturbances.

## 5. Challenge 1: Targeting Aerial Objects with UAVs

Challenge 1 focused on airspace safety and requires physical interaction with multiple aerial targets. 15 minutes were allowed for each trial. The challenge can be separated into two primary tasks:

**Balloon-Popping Task: Semi-stationary targets.** Pop five 60cm-diameter green balloons tethered to 2.5m poles with string and placed randomly throughout the arena.

**Ball-Catching Task: Moving target.** Capture and return a 15cm yellow foam ball hanging from a UAV flying throughout the arena in a figure-8 path at 8m/s.

Our approach consisted of an autonomous UAV for each task, both following a global search plan and deviating from that path when a target was identified. The system was tested in an outdoor arena-like space in Pittsburgh. Challenge 1 competition results are presented in Section 8.1.

## 5.1. Hardware

Both robots for Challenge 1 use the common platform described in Section 3 with lightweight attachments but no additional task-specific sensors.

### 5.1.1. Lightweight attachments

For the Balloon-Popping Task, we found the propellers effective at puncturing the thick-latex large balloons without destabilizing the vehicle, given safe servoing that prevented bodily contact with the pole. The Ball-Catching Task platform was modified as seen in Figure 3, with a custom carbon fiber rod-mounted net, and wire strung across the top to maintain tension and provide protection from the incoming target. This design is simpler and lighter than a robotic arm and, additionally, covers a large area and range of angles for the incoming ball, held magnetically to a carrier UAV. The DJI Center-of-Gravity Auto-Calibration Tool improved flight stability with the net mounted.

## 5.2. Control strategies and architecture

Figure 4 presents the state machine for Balloon-Popping Task. After takeoff, the UAV follows a global search plan and deviates when it detects a valid target. During global search, if the battery was low

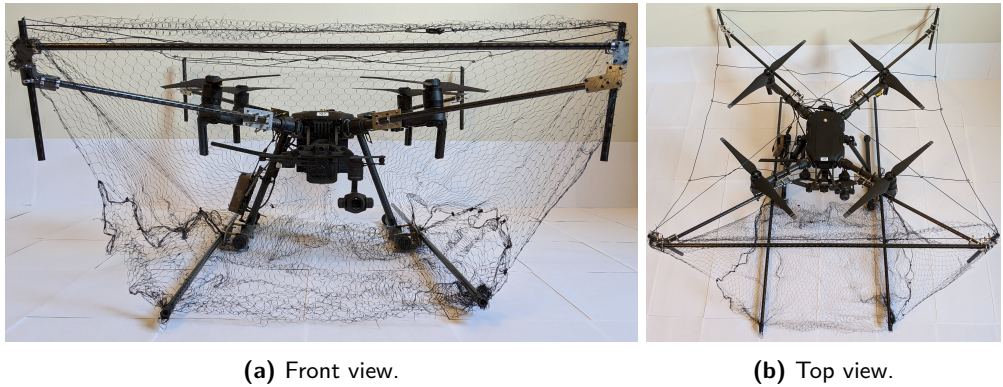

**(a)** Front view.  **(b)** Top view.

**Figure 3.** Challenge 1 Ball-Catching Task UAV with mounted ball-capture structure.

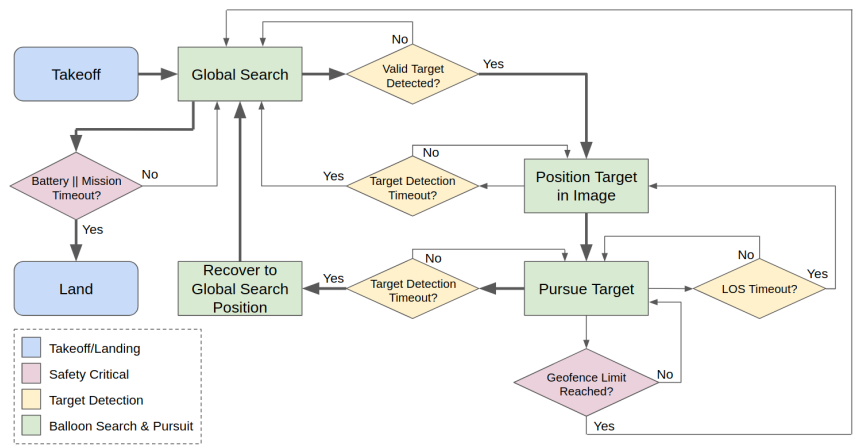

**Figure 4.** State machine for Challenge 1 Balloon-Popping Task. Bold arrows depict a standard flow of the system.

or if the internal mission clock neared the 15-min limit, then the robot landed. Figure 5 presents the state machine for Ball-Catching Task. Takeoff/Land, Safety Critical, and other subsystems are similar to those in the Balloon-Popping Task system. Once the ball is detected, the UAV exits the global search and enters a position-refinement loop where, upon each pass of the target, it attempts to get closer to an intercept point with the ball's path.

### 5.2.1. Global search and local interaction

A global search plan was used in both Challenge 1 tasks. For Balloon-Popping Task (Figure 6a), this path involved a forward pass and a shifted, backward pass lawnmower scan according to measured GPS coordinates of the arena. The sweep width (6m), altitude (2.7m), and velocity (2m/s) were tuned during competition. A valid target detection paused the global planner and began local interaction, which composes of the `Position Target in Image` and `Pursue Target` stages. In the first stage, the UAV positions itself to achieve a centered and upward line-of-sight (LOS) vector towards the target to ensure that the propellers hit the balloon in the second stage, where the UAV moves along the LOS until the target is lost. Detection timeouts are used to ignore false positive detections and trigger recovery behavior after pursuing a balloon. If a valid target detection remained but the LOS pursuit timed out (`LOS Timeout?`), then the UAV transitioned to pursue it as a new target, thus allowing multiple balloons to be popped when in close proximity.

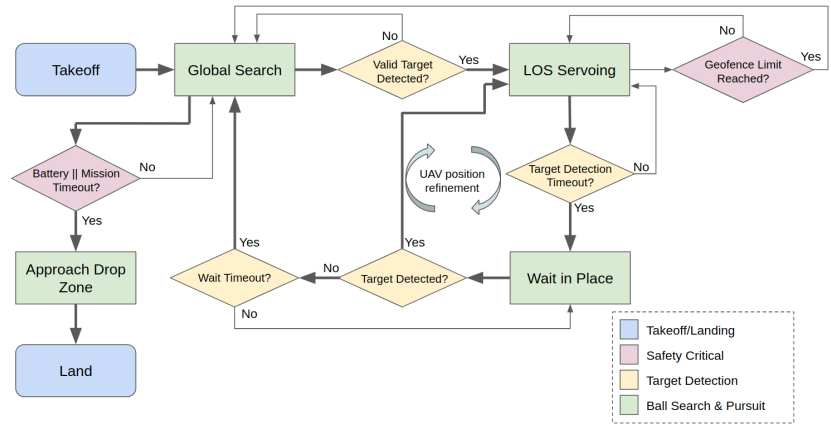

**Figure 5.** State machine for Challenge 1 Ball-Catching Task. Bold arrows depict a standard flow of the system. We label the loop that incrementally refines the UAV position for intercept with each pass of the ball.

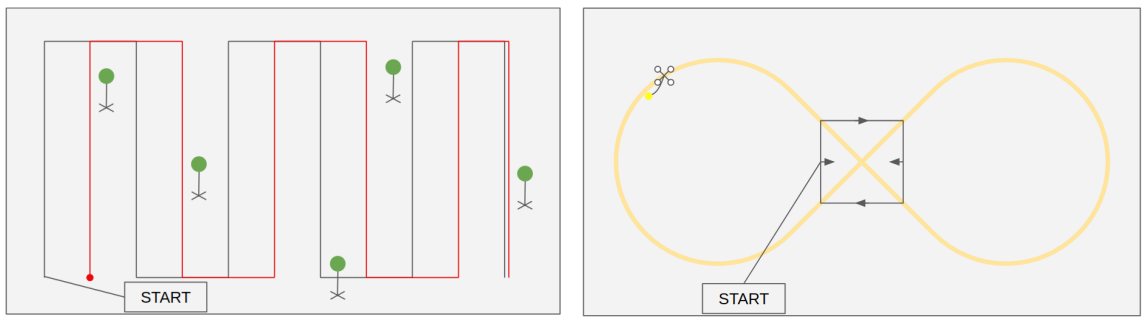

**(a)** Balloon-Popping Task global search. Black is the forward pass and red is the shifted reverse pass.

**(b)** Ball-Catching Task global search. Black lines and arrows indicate our UAV's path and heading, respectively. In yellow is the target's path.

**Figure 6.** Top-down illustrations of global search plans for Challenge 1 tasks.

For Ball-Catching Task (Figure 6b), the global planner produced a square trajectory at the center of the arena at two pre-set altitudes (10m and 12m), both higher than the balloons. The UAV orientation throughout the trajectory aligned with the long side of the arena, thereby facing the center of the target's figure-8 path, ensuring that the target is in view and in close proximity for the longest time possible. Once the target is identified, the UAV halts forward movement and moves laterally and vertically according to the LOS direction to intercept the ball. Once the target passes out of view, the UAV waits in place for the next pass, upon which it servos again; if it does not see the target for some `Wait Timeout?` (set to approximately two periods) then it reverts to the global plan. Figure 5 shows this position-refinement loop in the state machine flow. Notably, even if the ball is caught, the UAV lands only upon mission timeout. This was done simply because of the lack of a catch detection module during competition. The strategy here therefore does not prioritize the minimum possible mission time, but instead enables multiple attempts at catching the ball.

### 5.2.2. Outcome monitoring and recovery strategies
The Balloon-Popping Task state machine naturally encodes a recovery strategy into the mission structure. After target pursuit times out (`Target Detection Timeout?`), the UAV flies up, back, and down to the original point on the global search from which it detected the target. This strategy serves to either continue the global search or, if the balloon was not popped, re-identify the target

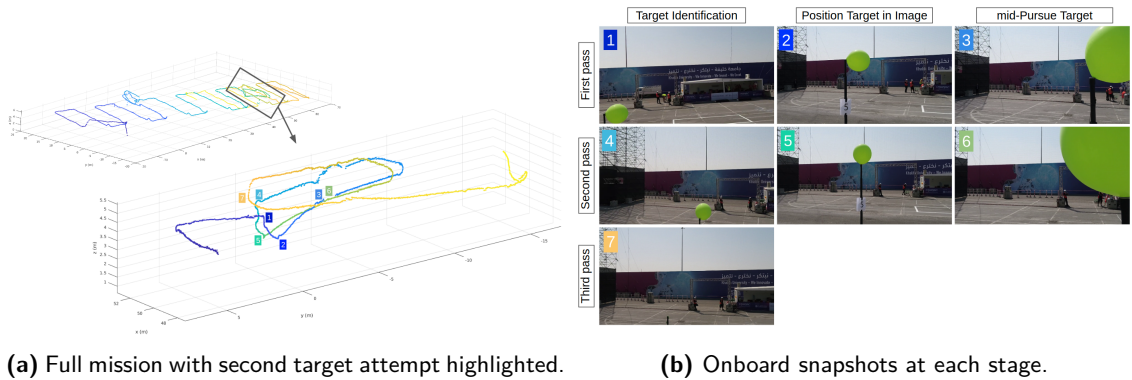

**(a)** Full mission with second target attempt highlighted.                 **(b)** Onboard snapshots at each stage.

**Figure 7.** Examination of competition Balloon-Popping trial run in which gimbal camera was off-center. In the full trial run view in 7a (top), it can be seen that the UAV proceeds on the lawnmower search path in the forward pass (blue to green) and then half of the reverse pass (orange to yellow). Two targets were attempted on this run, the second of which is analyzed in 7a (bottom), with corresponding numbered onboard snapshots shown in 7b. Key points are shown in the local interaction subtasks, showing the first and second passes, and the beginning of the third pass where the target is no longer observed (frame 7) after successful contact (frame 6).

and re-attempt the pursuit. We found that this simple Detect-Attempt-Recover-Re-Detect cycle, executed naturally by the flow of the state machine (indicated with bold arrows in Figure 4), was very reliable and provided robustness to environment variation and system failures. The balloons can sway and miss the UAV propellers, but re-attempts resulted in success even in high winds. Figure 7 depicts a competition trial in which the gimbal camera angle had a constant offset from the UAV body frame due to an error in setup. This misalignment causes servoing in the wrong direction since it was assumed during development that the robot body and gimbal frames are always aligned. The snapshots show the UAV missing the target on the first pass, but successfully popping the balloon on the second pass since wind moved the target just enough to make contact with the propellers. Thus, we demonstrate mission robustness even to system faults. Furthermore, the `LOS Timeout?` condition allows the UAV to re-position itself relative to the target, thereby re-initializing the pursuit action, if the target is still in view after some pursuit duration.

## 5.3. Vision pipeline

In this section we describe the full vision pipeline of the Challenge 1 systems, from target detection to interaction.

### 5.3.1. Robust detection in changing environments

Color thresholding and ellipse fitting were tried for detection of the distinct green balloon and yellow ball targets, but required extensive tuning for different environments. We therefore used a deep-learning approach for better generalization to the unknown competition scene. TinyYOLOv3 (Redmon and Farhadi, 2018), a single-shot object detection neural network, provided high accuracy and real-time inference (24-30fps) when paired with the Intel OpenVINO framework (Gorbachev et al., 2019) for CPU inference optimization compared to MobileNet-SSD (Howard et al., 2017), YOLOv3, and ResNet-50 (He et al., 2016). Hand-collected data of balloons and balls were collected to train two models starting from the ImageNet-pretrained weights. Data augmentation was used to increase the size and variation of the dataset, and the trained models worked well in both sunny and blizzard-like weather. Finally, a valid balloon-target detection requires the bounding box to be larger than some threshold, typically tuned for a upper detection range limit of ~4m; when there are multiple detections, the largest is considered the current target.

Though separate balloon and ball detection networks were trained, each network often triggered false positives on the opposite target. The balloon detector was never shown the higher-altitude ball, but the ball network frequently detected the balloons in the bottom of the image frame. Instead of relying on a tuned color discriminator which might suffer with lighting changes, and since the global search pattern is slightly lower than the ball's path ensuring that a detected ball would be in the top half of the image, we simply filter out all detections in the bottom 40% of the image to ignore the balloons.

### 5.3.2. Continuous visual feedback for precise interactions

Line-of-sight (LOS) guidance controlled local interaction with targets. This strategy is computationally cheap and requires only a monocular camera. The LOS vector $\mathbf{r_t}$ originates from the camera center and passes through the image frame at the pixel centroid of the target, pointing towards the target:

$$\mathbf{r_t} = \begin{bmatrix} \dfrac{u_t - c_x}{f_x} \\ \dfrac{v_t - c_y}{f_y} \\ 1 \end{bmatrix} \tag{1}$$

where the target's pixel centroid is represented by $(u_t, v_t)$; the camera principal point, $(c_x, c_y)$, and the focal lengths, $f_x$ and $f_y$, are obtained from the calibrated camera's intrinsic matrix. This vector was used for the `Position Target in Image` and `Pursue Target` stages for Balloon-Popping Task (Figure 4) and the `LOS Servoing` stage in Ball-Catching Task (Figure 5).

## 6. Challenge 2: Autonomous Block Pick and Place

Challenge 2 focuses on construction, requiring locating, transporting, and placing foam blocks of varying color (red, green, blue, orange) and mass (1-2kg) that have ferromagnetic white patches (Figure 8a). The UAVs were to place the blocks on an elevated `W`-shaped structure (1.7m tall) with yellow channels (Figure 8b). Block pickup and placement required precise and stable interactions, necessitating the development of a tuned visual detection and servoing pipeline. Though Challenge 2 included UGV tasks, we do not address those in this paper. Challenge 2 competition results are presented in Section 8.2.

### 6.1. Hardware

We added sensors to the common UAV platform for detecting and measuring both the blocks and placement structure. A custom end-effector fitted with electro-permanent magnets was used to pick up the blocks.

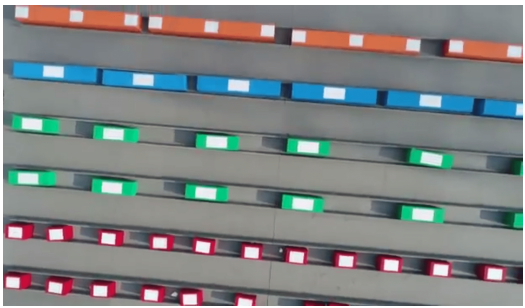

**(a)** Block pickup site.

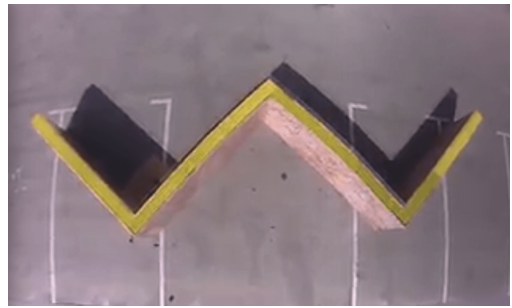

**(b)** Block placement structure site.

**Figure 8.** Challenge 2 arena sites of interest from the UAV's onboard gimbal camera.

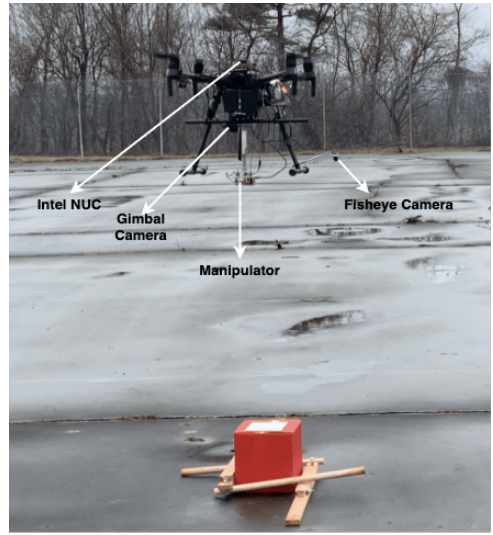

**Figure 9.** Challenge 2 UAV with labeled hardware during a block pickup test in Pittsburgh.

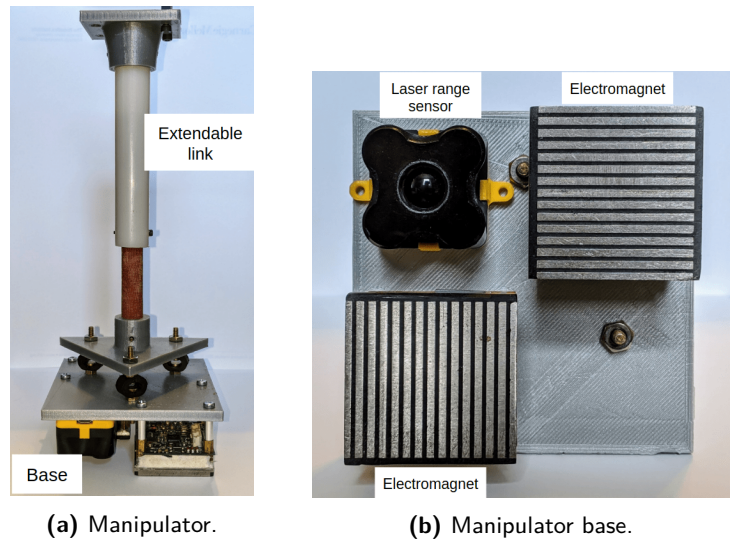

**(a)** Manipulator.  **(b)** Manipulator base.

**Figure 10.** Manipulator for picking up blocks with electromagnets.

### 6.1.1. Task-specific sensors

During block pickup, the gimbal camera points downward and is used for block detection and visual servoing. A short-range distance sensor (TeraRanger Evo 3m Specifications, 2020) was used on the base of the block pickup manipulator (Figure 10b) to detect block contact. After a block has been successfully picked up, the field-of-view (FOV) of the gimbal camera is obstructed by the block, so a fisheye lens camera attached to the UAV's leg (Figure 9) was used to detect the large placement structure and visually servo to the exact dropoff location.

### 6.1.2. Lightweight attachments

A simple and lightweight extendable manipulator with a magnetic base, composed of 3-D printed plates and a plastic/cardboard link, was used to pick up blocks (Figure 10). We used two OpenGrab EPM v3 (NicaDrone OpenGrab EPM Datasheet v3, 2020), similar to (Bähnemann et al., 2019) and (Castaño et al., 2019), which are lightweight electro-permanent magnets with minimal power consumption. Three vibration dampers (black rings in Figure 10a) increased compliance and protected against impact.

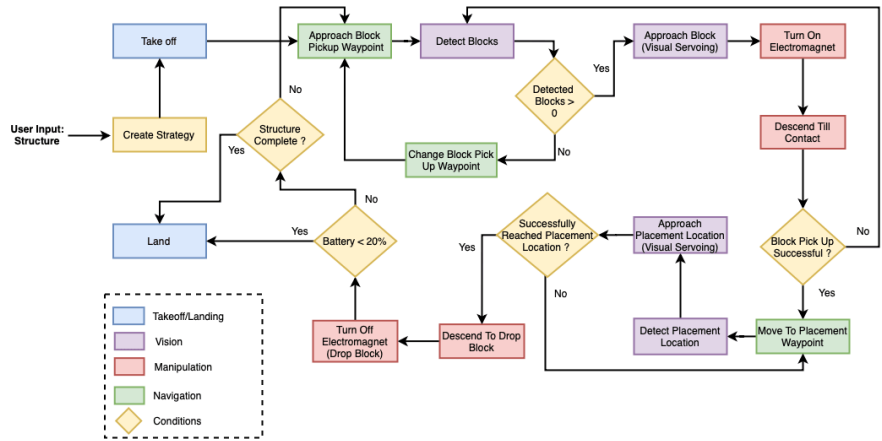

**Figure 11.** Challenge 2 state machine for block pickup and placement.

## 6.2. Control strategies and architecture

Figure 11 presents the state machine for this challenge. The strategy originally depended on a pre-specified brick configuration, designated as `User Input`, but was later simplified to placing red blocks relative to one side of the placement structure. After a block was placed, the visual servoing goal coordinates were adjusted to place another red block adjacent to the previous one. Various outcome monitoring triggered transitions in the state machine.

### 6.2.1. Global search and local interaction

Global search for both block pickup and placement begins with navigation to a measured GPS waypoint at altitude above the two sites (3.5m and 6m, respectively). If no blocks of the chosen color are found, then the UAV moves to a nearby point sampled from a Gaussian to continue the search. Once the block or placement structure is identified, visual servoing aligns the manipulator above the target location. After convergence, the UAV drops in altitude and either (for pickup) uses the altitude sensor and manipulator range sensor to detect block pickup, or (for placement) drops the block, then increases altitude. The mission terminates once the battery is low.

### 6.2.2. Outcome monitoring and recovery strategies

Recovery behaviors were possible at different stages of block pickup. If block detection fails then the UAV recovers to the original global search behavior. If the UAV loses track of the block patch's four corners during visual servoing, it recovers to a height of 2.5m above the last-seen block's location and re-attempts the vision pipeline. If the UAV descends too far during block pickup descent (descent velocity is controlled in a closed loop using the laser altimeter) thus indicating that the manipulator missed the block altogether, or the manipulator's short-range sensor does not detect an attached block after increasing in altitude after a pickup attempt, then the UAV recovers to a height of 2.5m to re-attempt pickup. This recovery and re-attempt strategy enabled multiple safe attempts of block pickup during competition, as described in our results.

## 6.3. Vision pipeline

The vision pipeline is responsible for object detection and tracking certain keypoints accurately enough for generating control input. Namely, during block pickup the UAV uses the gimbal camera

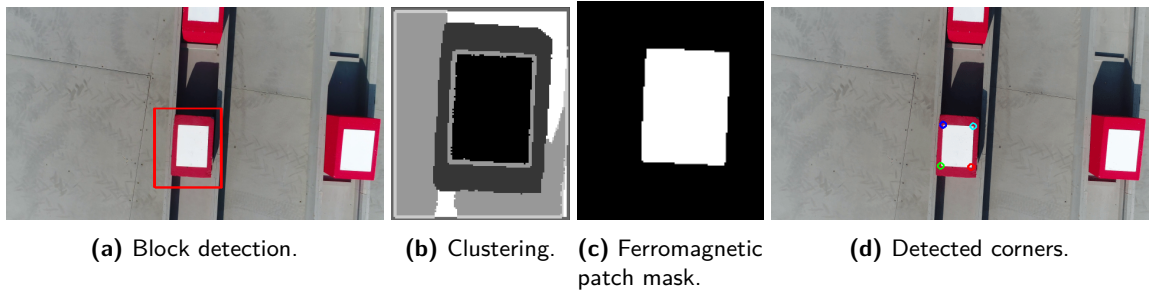

**(a)** Block detection.  **(b)** Clustering.  **(c)** Ferromagnetic patch mask.  **(d)** Detected corners.

**Figure 12.** Block corners detection pipeline.

(facing downward) to track the four corners of a block's white ferromagnetic patch, and during block placement the UAV uses the fisheye camera to track two corners of one side of the placement structure. This pipeline had to run in real-time and be easily tunable for new environments.

### 6.3.1. Robust detection in changing environments

**Block and metal patch detection**
Similar to Challenge 1 (Section 5.3.1), TinyYOLOv3 paired with the OpenVINO inference engine was used for generalizable 4-class block detection at real-time (28fps observed). The block of the desired color with the largest bounding box is chosen as the pickup target, heuristically finding the closest block. A median flow tracker (Kalal et al., 2010) was chosen to track the target block because of its high framerate and good failure detection.

Next, the bounding box is segmented using k-means clustering (Kanungo et al., 2002) in HSV space (Sural et al., 2002) (Figure 12b) and the ferromagnetic patch is extracted by taking the cluster closest to a calibrated color value from a patch (Figure 12c). A rectangle is then fitted to the patch mask and its sorted four corners (Figure 12d) are fed to the visual servoing control algorithm. Failure detection in this pipeline consists of comparing the length-to-width ratio of the fitted rectangle (units of pixel:pixel) to the known length-to-width ratio of the ferromagnetic patch (units of meter:meter). This ratio is scale invariant since we assume the gimbal camera is viewing at an angle perpendicular to the block face and the block is on a flat surface, such that the perspective 3-D projection into the image frame reduces to a similarity transform. This check helped detect tracker drift or selection of the wrong k-means cluster, which would trigger a restart of the vision pipeline.

**Placement structure detection**
For block placement, we used the two end corners of one side of the W-structure. The larger target size and different background features made it difficult to tune the number of clusters for k-means. Instead, a binary segmentation mask of the structure was obtained from training a logistic regression model on HSV pixels, which worked well because of the high structure-to-background contrast. Not having visual data of the structure prior to competition, we chose this approach since it was quickly tunable with very little onsite data. A polygon was fitted to the mask and all corners identified, with the top-most pair in the image targeted since we always approach the structure from the same perspective. A median flow tracker around these corners is re-initialized every 50 frames upon descent since the increase in details seen on the structure can cause it to fail. This process is shown in Figure 13.

### 6.3.2. Continuous visual feedback for precise interactions

Image-based visual servoing (IBVS) provides a mapping between pixel error in feature points to body frame velocities that would minimize the error (Sanderson, 1980). IBVS was used for closed-loop control to a selected ferromagnetic patch on a block (using four target corners) and to one side of the placement structure (using two target corners). This method is more robust to errors in camera

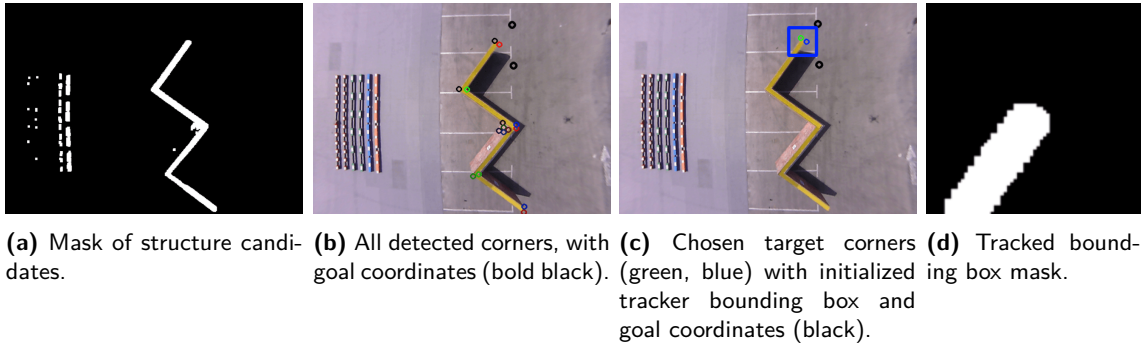

**(a)** Mask of structure candidates.

**(b)** All detected corners, with goal coordinates (bold black).

**(c)** Chosen target corners (green, blue) with initialized tracker bounding box and goal coordinates (black).

**(d)** Tracked bounding box mask.

**Figure 13.** Placement structure corners detection pipeline.

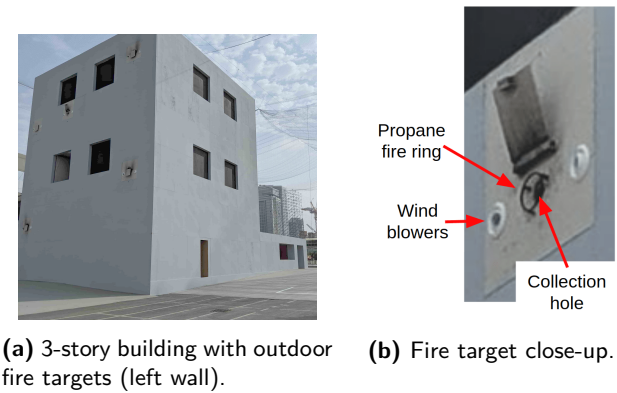

**(a)** 3-story building with outdoor fire targets (left wall).

**(b)** Fire target close-up.

**Figure 14.** Challenge 3 arena building and outdoor building fire target.

calibration and noisy measurements than a full pose estimator (like perspective n-point projection (Fischler and Bolles, 1981)). The details of our approach can be found in Appendix C.

## 7. Challenge 3: Autonomous Firefighting

Challenge 3 focuses on firefighting and requires dispensing water into holes on a fire target or flame-resembling target. We chose to use a UAV to target the building's outside-wall, propane-fueled targets, shown in Figure 14. Though Challenge 3 included UGV tasks, we do not address those in this paper. Challenge 3 competition results are presented in Section 8.3.

### 7.1. Hardware

The common platform was modified with sensors and attachments to locate the fire target and then dispense water onto it efficiently.

#### 7.1.1. Task-specific sensors

For thermal detection, we use a FLIR BOSON 320 thermal camera (FLIR BOSON Thermal Core Specifications, 2019). To enable horizontal wall following at a fixed distance, we use a Terabee EVO 60m laser range sensor (TeraRanger Evo 60m Specifications, 2020) mounted to point directly forward at the wall face. These are both seen in the front sensory unit in Figure 15b.

#### 7.1.2. Lightweight attachments

For this task, we needed a lightweight, high-capacity water pump suitable as a UAV payload. Pressurized tanks are typically too heavy and difficult to refill. We instead designed a custom

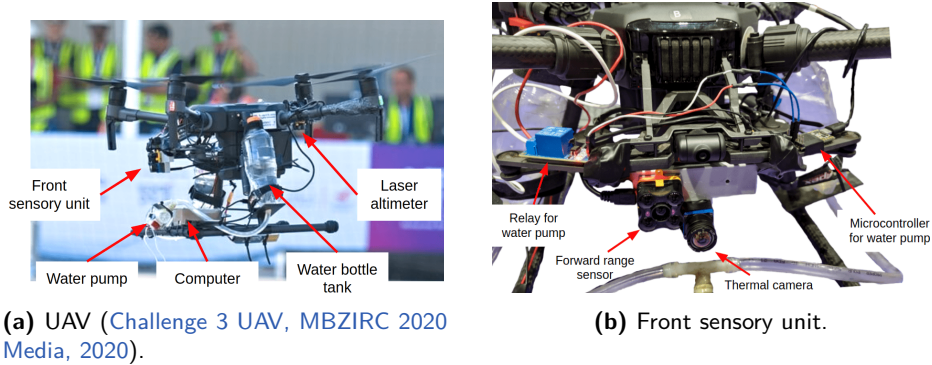

**(a)** UAV (Challenge 3 UAV, MBZIRC 2020 Media, 2020).

**(b)** Front sensory unit.

**Figure 15.** Challenge 3 UAV with annotated sensors and attachments.

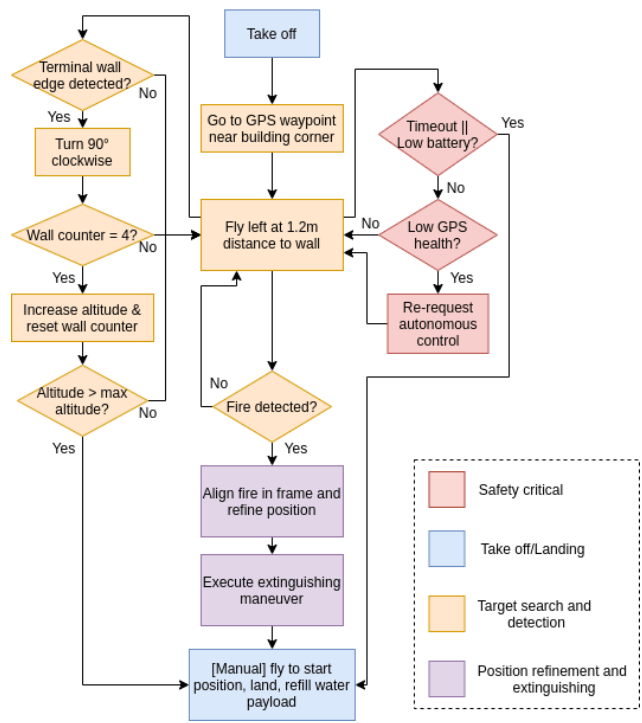

**Figure 16.** Challenge 3 firefighting state machine. Note that `Low GPS health?`→`Re-request autonomous control` was included to deal with fluctuations in GPS signal around the building.

solution composed of two plastic water bottles connected to a vehicle's windshield washer pump, activated with an electromagnetic relay and a Teensy 3.6 microcontroller (Teensy 3.6 Datasheet, 2015). This design had a maximum capacity of 1.5L, and is shown in Figure 15a. Since the water tanks are standard screw-cap bottles, we were able to refill them quickly with just a funnel, leading to eight full mission cycles (each with a refill) in a single 15-minute competition trial.

### 7.2. Control strategies and architecture

Challenge 3 presented various indoor and outdoor fire targets, including real propane-fueled thermal targets and flame-like visual, non-thermal targets. Given the 15-minute time limit, our strategy consisted of only tackling the multiple fire targets on the external wall of the building (see state machine in Figure 16). The UAV scans the building's external wall for a thermal target and executes local behavior to maximize the water volume collected in the target collection hole. Because of the difficulty in accurately monitoring the water entering the target's hole, and the limited onboard water payload, there was no outcome monitoring and recovery strategy for this challenge. After extinguishing, the UAV can automatically land for water refilling or wait for manual takeover to fly it back to the start position for water refilling.

### 7.2.1. Global search and local interaction

The UAV starts on the ground, directly facing the building wall. The global search plan begins with a GPS waypoint 2m forward and to the left at an altitude pre-set to target a specific floor's fires, turned clockwise 90° to pre-align with the relevant wall face for initializing the wall following algorithm. The UAV then proceeds to fly left, parallel to the wall, at a distance of 1.2m controlled with the forward range sensor. The search plan consists of wall-following spirals around the building at increasing altitudes (to tackle fires on different levels), though in competition the UAV always either found a target on the first wall pass or suffered a GPS signal failure. We assumed that each wall had a single starting and ending edge; a median filter was used to reliably detect these wall edges with just the forward 1-D range sensor. This filter was used to avoid open windows and also had a minimum distance traveled requirement per wall side (based on manually-taken measurements) for executing turns to avoid premature false positives of wall terminal edges. The wall following algorithm can be found in supplementary material on our website.

## 7.3. Vision pipeline

The vision pipeline for this system is responsible for identifying a thermal target and issuing control commands to position the water pump nozzle relative to it. For this challenge a robust, generalizable detector was not necessary since the fire targets were easily distinguishable from the background wall face. A thermal image pixel value threshold, where pixels are a function of the detected heat, was easily tuned and resulted in a binary mask in the image. We surround all connected components with a bounding box, eliminate those that are very small, and are left with candidate fires that we use for maneuvering (Figure 17b). During testing it was found that this can be tuned to avoid other warm objects, such as humans, in the scene.

### 7.3.1. Continuous visual feedback for precise interactions

After finding a fire in the thermal image frame, we use camera geometry and the forward range sensor (assuming the UAV is facing the wall) to find the approximate 3-D location of the target relative to the UAV. First, the UAV moves to center the target in the image frame via proportional gains on pixel error. Once this converges, the previously-calculated relative 3-D position of the fire is used to refine the position further. Visual servoing was not used since we did not know if the

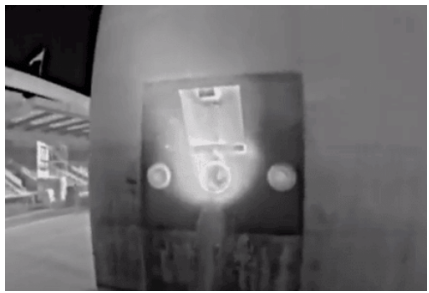

**(a)** Thermal image of water hitting the fire, just below the opening (water stream shown in lighter grey).

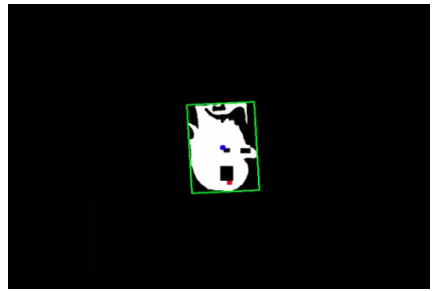

**(b)** Segmented thermal image with fire enclosed in green.

**Figure 17.** Fire detection during competition trial.

**Table 1.** Success and failure tally for Balloon-Popping Task subtasks across all competition trials

| Subtask | Success count | Success rate | Failure cause | Count |
|---|---|---|---|---|
| Target identification | 24/26 | 92% | Search pattern far from target | 2 |
| Pop sequence | 15/25 | 60% | High-rate turn when detected | 6 |
| | | | Gimbal off-center | 2 |
| | | | Propeller downdraft | 2 |
| Recover & pop | 4/4 | 100% | N/A | N/A |

target's corners would be easily detectable in the thermal image, and wanted to avoid adding an additional RGB camera. Once in position, the UAV performs a robust action to increase the chance of water entering the collection opening since we do not have a model of the water pump or the distance and spread of the spray, and since visual feedback measuring contact of the water is very challenging. This robust action is composed of flight in a ×-pattern hitting the corners of a 0.25m × 0.25m square.

# 8. Competition Results

We present all results from days containing successfully-started competition trials, with discussion describing key takeaways and immediate improvements. Omitted are competition trials during which irrelevant problems occurred (e.g., no GPS lock, hardware failures).

## 8.1. Challenge 1

### Balloon-Popping Task

The Balloon-Popping Task results are tabulated in Table 1 combining every trial on three competition days, including both Challenge 1-specific Grand Challenge trials. To keep results clear, data from one trial is excluded wherein bad system startup had caused multiple failures.

The *Target identification* subtask was defined by the UAV successfully registering a balloon as a valid target when at the closest point on the global search plan, denoted by the `Valid Target Detected?` condition in Figure 4. In both failure cases, the balloons were sufficiently far from the global search path and appeared too small for the minimum visual size requirement. This was largely a failure in correctly setting the global search path.

The *Pop sequence* subtask was defined as successfully passing through the states `Position Target in Image` and `Pursue Target`, resulting in a pop on the first attempt. The predominant cause of failure here was detection of a balloon during a turn; turns were tuned to be executed quickly for efficient coverage of the arena, but when a balloon was detected during a turn the UAV often lost sight of it by the time it entered the adjustment stage. Figure 18 features two runs of the UAV's position during a competition trial in which one balloon near the global path's edge was missed during high-rate turns. Although the UAV would turn in the target's direction, it often did not regain sight of it before the `Target Detection Timeout?` condition was triggered. This likely could have been solved by increasing the detection timeout or adjusting the turning behavior, but it was not successfully fixed during competition.

The *Recover & pop* subtask was defined as a successful cycle through the states `Recover to Global Search Position` and `Global Search`, and the subtasks *Target identification* and *Pop sequence*, resulting in a pop after a failed attempt. This behavior was triggered on the four occasions that *Pop sequence* failed due to off-center gimbal camera angle or propeller downdraft. *Pop sequence* failures due to high-rate turns occurred in the `Position Target in Image` state so recovery was not triggered. The UAV was successful in all four recovery-and-pop attempts.

### Ball-Catching Task

During competition trials, the UAV repeatedly found the target on the global search plan. However, each time, the UAV position-refinement loop was started at a point where the ball was approaching

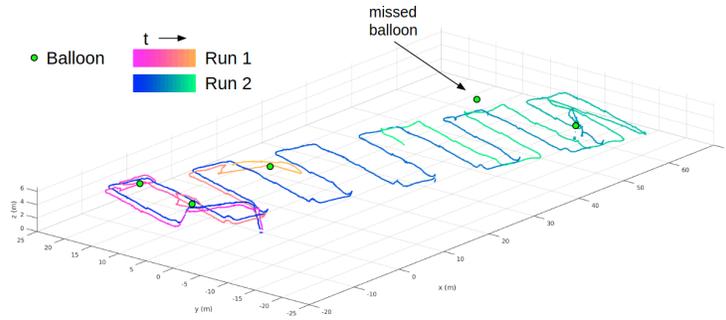

**Figure 18.** Two runs from the first day of Challenge 1 competition. 4/5 balloons were successfully popped over two runs. The missed balloon was only detected during high-rate turns.

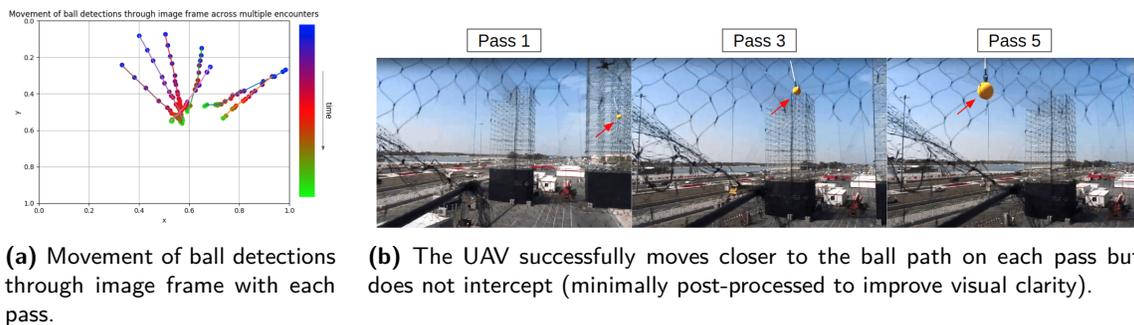

**(a)** Movement of ball detections through image frame with each pass.

**(b)** The UAV successfully moves closer to the ball path on each pass but does not intercept (minimally post-processed to improve visual clarity).

**Figure 19.** UAV position-refinement with each pass of the ball during competition. The ball is moving away from the UAV.

from behind and passed over the UAV; the UAV then detected the target as it moved away. While just touching the ball would have earned points, this circumstance induces a geometry in the camera frame that produces a vanishing LOS vector in the vertical and lateral directions. Since this vector controlled the UAV's position refinement, it made it impossible to converge to the ball's path. Figure 19a plots the ball's motion in the image frame over seven passes of the ball during the UAV's position-refinement loop. The target starts near the edges of the frame, producing a large LOS vector guiding the UAV towards the figure-8 path for intercept. However, as the target moves forward and away from the camera, it nears the center of the image frame and sometimes even issues LOS vectors opposite in direction from the initial part of the pass. Because of this, the UAV was able to get very close to but never successfully intercept the target, as seen in Figure 19b. Oscillations in UAV position were observed as the ball moved downward in the image frame, sometimes below the image center causing a downward LOS vector; however, this was limited since target detections in the bottom 40% of the image were ignored.

### Discussion

The results on the Balloon-Popping Task lead to a couple of key conclusions: ($i$) simple LOS guidance works when targeting semi-stationary objects and paired with an effective global search plan, and ($ii$) encoding simple checks and recovery behavior into the system, such as the recovery to the original search position, added robustness that overcame certain system faults and helped ensure mission success.

The ball-catching strategy failure was largely an oversight in development caused by limited simulation and real-world testing prior to competition, due to logistical challenges. We performed a follow-up simulation analysis, the details of which can be found in the supplementary material on our website. From the competition results and this simulation analysis, we can arrive at the following conclusion: targeting moving aerial objects with monocular vision requires a reactive guidance law. While simple LOS guidance worked for targeting semi-stationary objects, efforts in adapting this guidance for the Ball-Catching Task made clear that the task requires a richer sensor input, possibly enabling 3-D localization of the ball for trajectory estimation. With just monocular RGB, a guidance law such as proportional navigation might be adapted for a quadrotor and in simulation it was found to work well when in conjunction with our existing search-locate-refine strategy.

**Table 2.** Success and failure tally for Challenge 2 subtasks on the second day of competition trials

| Subtask | Success count | Success rate | Failure cause | Count |
|---|---|---|---|---|
| Block detection | 12/13 | 92% | Malfunctioning camera | 1 |
| Visual servoing convergence | 12/12 | 100% | N/A | N/A |
| Single-attempt block pickup | 6/12 | 50% | Large drift in final descent | 4 |
| | | | Small drift (partial contact) | 2 |
| Block placement | 1/6 | 17% | Incorrect desired corners | 2 |
| | | | Incorrect mask generation | 3 |

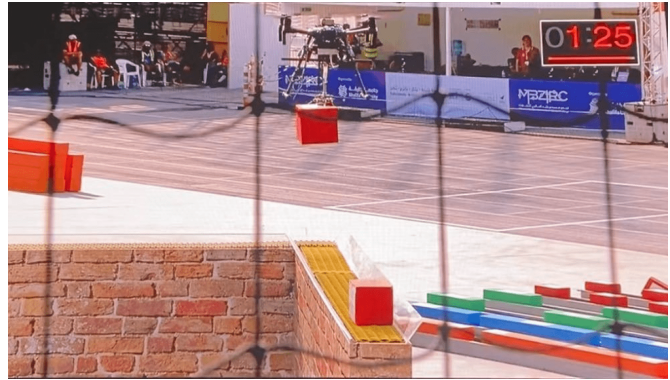

**Figure 20.** The UAV descending for another block placement attempt after a successful placement on the second day of competition trials.

### 8.2. Challenge 2

Missions from the Challenge 2 second competition day are split into key subtasks for analysis (hardware faults on other competition days precluded trial runs) and tallied in Table 2. A successful *Block detection* correctly identified a full block of the requested color; this failed once due to a malfunctioning camera. A successful *Visual servoing convergence* controlled the UAV camera position relative to the block corners until convergence over the block; this suffered no failures. A successful *Single-attempt block pickup* achieved stable contact between the UAV manipulator's electromagnets and the block's ferromagnetic patch in an isolated pickup attempt, such that the pickup was maintained throughout UAV flight to the placement location. The final 0.7m of descent to the block is open-loop since the gimbal camera FOV is not wide enough to capture the block at close range. The most common failure case was lateral drift during this open-loop portion, caused by external factors like wind. We tabulate large drift, in which the magnetic end effector did not contact the ferromagnetic patch at all, and small drift, in which partial magnetic contact was achieved but dropped the block upon ascent. A successful *Block placement* placed the block on the structure (Figure 20). Two failure cases were observed for this subtask, both in the vision pipeline; either the detected corner pair of the structure's end was shifted, causing repeated block placements to the right of the structure, or a faulty structure mask resulted in tracking the wrong part of the structure.

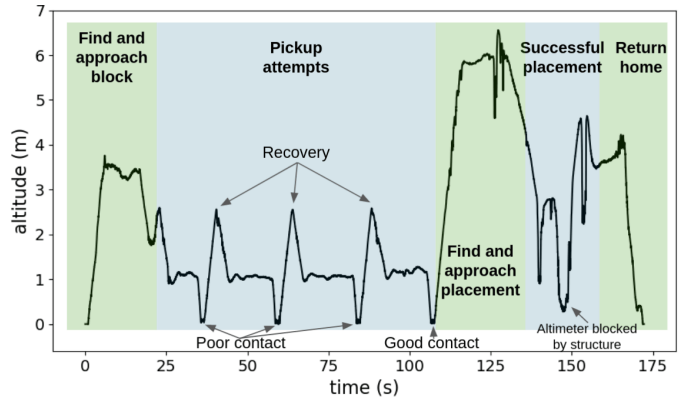

**Figure 21.** Altitude of the UAV for a takeoff-block pickup-block placement-land mission.

**Table 3.** Success and failure tally for Challenge 3 subtasks across all competition trials

| Subtask | Success count | Success rate | Failure cause | Count |
|---|---|---|---|---|
| Wall following | 19/24 | 79% | Faulty forward range sensor readings | 5 |
| Fire detection | 17/23 | 74% | False negative | 2 |
|  |  |  | False positive | 4 |
| Fire extinguishing | 5/21 | 24% | Inaccurate fire localization | 2 |
|  |  |  | Water supply too low | 10 |
|  |  |  | Stuck water pump stopper | 4 |

### Discussion

The open-loop final 0.7m descent for block pickup resulted in a low *Single-attempt block pickup* success rate, but our recovery-and-re-attempt behavior resulted in a successful pickup for every mission attempted in competition (except one in which the gimbal camera malfunctioned). A sample mission's UAV altitude is plotted and annotated in Figure 21. While the gimbal camera improved visual servo control stability during block pickup as the UAV moved, we might have made the final 0.7m descent stage closed-loop by utilizing the wider-FOV fisheye camera. RTK-GPS, though intentionally avoided, or visual odometry (tracking keypoints and correcting for lateral positional drift) during the final descent might have also improved the *Single-attempt block pickup* success rate. Use of the fisheye camera for block search might have also enabled the UAV to operate at a lower altitude when searching for blocks, saving battery life and reducing mission time.

### 8.3. Challenge 3

Challenge 3 results are compiled from the trials runs on all three competition days and tallied in Table 3. A successful *Wall following* followed the wall at a safe distance and correctly turned corners without entering any windows. A successful *Fire detection* registered true positive fire detections in the vision pipeline; two false positives triggered the extinguishing system, caused by recently deactivated targets that were still hot from the competition trial immediately prior, and two other false positives were caused by thermal detections seen through windows, causing the UAV to enter the windows (explained in detail below). We consider a successful *Fire extinguishing* as dispensing water on the metal target, in or near the 12cm opening. Twice, the UAV dispensed water too far from the collection hole due to inaccurate fire localization in the vision pipeline. The most common failure occurred during the Grand Challenge, when resets were limited due to the concurrently-running Challenges 1 and 2, barring team members from entering the arena and refilling the UAV's water supply. This caused the UAV to continue finding fire targets but not dispense water. On four other

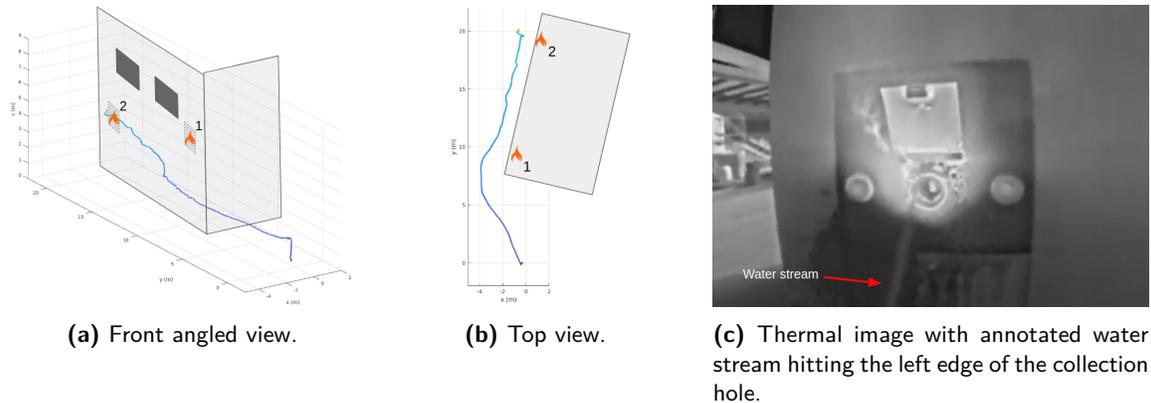

**(a)** Front angled view.

**(b)** Top view.

**(c)** Thermal image with annotated water stream hitting the left edge of the collection hole.

**Figure 22.** Odometry position and thermal image during the most successful run, in which we dispensed the most water autonomously into the target of all teams (target 2).

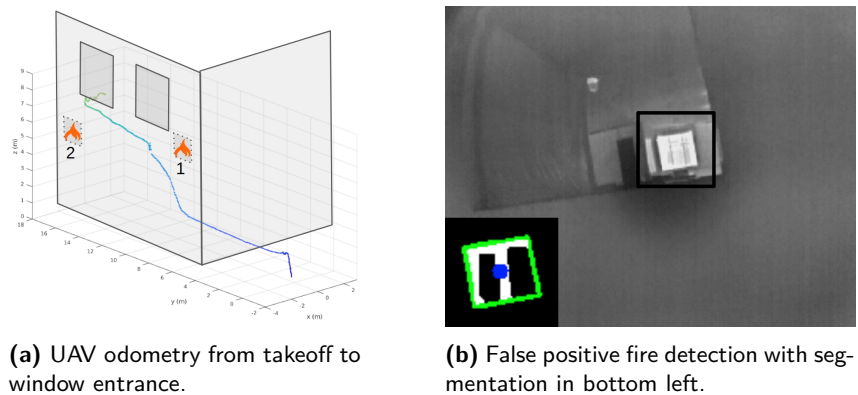

**(a)** UAV odometry from takeoff to window entrance.

**(b)** False positive fire detection with segmentation in bottom left.

**Figure 23.** False positive fire detection failure case causing the UAV to enter a window.

occasions, the water pump was unable to eject the leakage stopper, presumably due to either human error during resets or uneven heating of the plastic nozzle components in the strong sun.

In the Grand Challenge, we scored the most points in autonomous mode by a UAV compared to all other teams. Our UAV dispensed 29mL of water into the target's collection hole while the target's wind thrusters were enabled, adding turbulence to the conditions. Figure 22 shows the odometry of the UAV during this trial and also shows a thermal camera image during the fire extinguishing maneuver.

**Window entering failure case**

Though not part of the mission strategy, the UAV twice attempted to enter the building through a window and in one instance fully entered before the safety pilot took control and landed safely inside. The odometry position during this mission is shown in Figure 23a; fire target 1 was a false negative in the detection pipeline, but a false positive detection was registered looking through the window on what appears to be the window on the building's far side (Figure 23b). Because this was consistently detected and the contiguous contours were large enough, the UAV transitioned to position itself relative to this target, moving forward through the window to get closer to it before the safety pilot takeover. This might have been avoided with a simple upper limit on the target distance; however, an improved fire detection module incorporating flame classification might have reduced both the false positives and negatives.

**Discussion**
The global search and fire detection strategies worked well, almost always finding a target. However, the wall following method might have been improved simply by using more than one range sensor set at an angle to each other, a low-resolution depth camera (popular RGB-D cameras were avoided, however, due to their common USB complications), or a RADAR sensor (which has the additional benefit of detecting glass in a more modern structure). The fire detection strategy might be improved by using a histogram-based threshold, or – to avoid the residually hot targets or window-entering case – flame-detection via background subtraction. Since the current water dispensing mechanism has a narrow stream, it might be possible to aim the UAV with feedback from the thermal imagery by detecting sudden drops in temperature caused by the water's contact with the target. However, this was not pursued, since such an algorithm's efficacy would depend on the thermal properties of the competition-spec target. Finally, while the × trajectory resulted in success compared to other teams, there was occasionally some drift during the maneuver; it might be possible to minimize this by executing the action relative to the visually localized fire, i.e., having some form of visual feedback from the fire's pixel centroid.

## 9. Lessons Learned

Throughout the development of these rapidly deployable, modular, mission-robust systems we learned many valuable lessons that may be useful for other field robotics researchers.

**An autonomous deployment sequence drastically improves reset efficiency**
Our rapid-deployment pipeline avoids reliance on centralized communication infrastructure and notably allowed multiple productive, back-to-back runs during the time-limited competition trials. For example, in the Grand Challenge, between the frequent resets called for the concurrent Challenges 2 and 3, we were able to adjust Challenge 1 mission parameters and complete seven mission runs for the Balloon-Popping Task in under 20 minutes. In a single 15-minute Challenge 3 trial, we successfully achieved eight mission runs complete with water tank refills.

**Encoding checks in the mission structure adds robustness (and can overcome system faults)**
As detailed in Figure 7, in a Balloon-Popping Task trial, the gimbal camera heading was incorrect, resulting in failed attempts on two balloons, but the Detect-Attempt-Recover-Re-Detect mission structure successfully popped both balloons. In Challenge 2 (Figure 21), pickup checks triggered multiple block-pickup attempts, resulting, finally, in successful pickups.

**Developing an adaptive system increases field-testing efficiency**
Field testing in Pittsburgh required frequent tuning of the vision systems for sunny, overcast, and snowy weather. To obviate this step, we developed an adaptive vision pipeline for Challenge 2 that re-calculated the logistic-regression parameters for placement-structure detection from a single flight over the test site. Though it was not needed in the unchanging setting of the Abu Dhabi arena, it greatly improved our testing efficiency. See Appendix B for details.

**Designing reusable algorithms accelerates development**
Allocating sufficient development effort to a particular subtask's algorithm so that it is easily adaptable to another subtask can lower development time and improve algorithm generalizability. In Challenge 2, for example, developing the vision pipeline and servoing control for block pickup took significant time and effort, but both were easily modified for use in block placement (Figures 12 and 13). This strategy accelerated development, and algorithm improvements could be applied to both subtasks automatically.

**Robust actions can compensate for limited feedback control**

In Challenge 3, it was nearly impossible to estimate in real time how much water entered the target's 12cm opening, thus limiting the capacity for feedback control during the operation. The lightweight pump's inconsistency in range and spread of the spray further exacerbated this problem. To overcome this, the UAV performs a ×-shape trajectory maneuver during extinguishing, which allowed us to get some water in or very close to the target opening on every extinguishing attempt. Similarly, in Challenge 1, adjusting for balloon sway in the wind is difficult, so the UAV uses a slightly upwards trajectory for approach. This minimizes downdraft effects, and also maximizes the chances of the propellers contacting the target by aiming near the bottom of the balloon, which is the part of the balloon moving least when swaying.

**Minimize sensor complexity and use competition-like testing to mitigate hardware failures**

Our systems notably have only a few, simple sensors (RGB/thermal cameras, 1-D laser range sensors) to accomplish the competition tasks, which configuration we found more manageable than a large sensor suite when debugging system failures with a limited team size. For example, the simpler USB fisheye camera never failed, but the gimbal camera with advanced features failed many times, requiring additional software checks, thus complicating competition trials. Competition-like testing is also very important: some of our sensors, including the range sensors, under-performed in the strong Abu Dhabi sun and heat, limiting our altitude and distance estimation.

## 10. Conclusion

In this work we present completely autonomous, aerial robotic systems for tackling the problems of airspace safety, construction, and firefighting posed by MBZIRC 2020. The robots developed for each challenge maintain a theme of simple but robust hardware choices and algorithms – ranging from object detection to mission planning – that proved robust to changes in scenery, lapses in hardware performance, or tightening time constraints. Furthermore, these systems share a common hardware/software backbone, attesting to the modularity of our approach. They performed well in the windy, sunny, hot environments of Abu Dhabi, after being developed in the winter weather conditions in Pittsburgh. We describe development challenges from field testing and the failures discovered onsite in competition, leading to several lessons learned in encoding robustness into the mission structure, extending vision algorithms to new environments, and avoiding sensor failures. We believe our efforts can be useful to other field roboticists developing such systems.

In the final competition rankings, CMU Team Tartan ranked 8th in Challenge 1, 4th in Challenge 2, and 7th in the Grand Challenge (Challenge 3-specific trials suffered from GPS faults, resolved with the GPS check described in Figure 16). During the Grand Challenge, we tied for the 3rd best Challenge 1 score, faced persistent gimbal camera faults in Challenge 2, and got the 4th best Challenge 3 score. Notable achievements by our team throughout the three-day competition include popping all five balloon targets, being one of a few teams to get within 1m of the flying ball target, being one of only three teams that successfully placed a block with a UAV, and dispensing the most water into a fire target with a UAV (only two other teams were able to do so at all), all autonomously. Many of these tasks require centimeter-level accuracy to be completed successfully, attesting to the strength of our vision pipelines and mission structure concepts.

Future directions of this work include multi-robot planning and interaction to complete each task more efficiently. For example, the Balloon-Popping Task and the firefighting task could be accomplished with multiple UAVs and coordinated task-scheduling to maximize efficiency. Using multiple UAVs for block stacking would not only increase efficiency but also introduce the possibility of placing the larger, heavier blocks with collaborative pick-and-place. Another extension may involve pairing a UAV with a robot that has a different set of strengths, such as a ground vehicle (UGV). In firefighting, a UGV could help refill a UAV's water tanks during a mission and could also douse fires in tight, cluttered situations not suitable for an aerial vehicle.

## Acknowledgements

This work was supported by the Mohamed Bin Zayed International Robotics Challenge 2020 and the National Science Foundation, grant no. DGE1745016. We would like to acknowledge the hard work of Noah LaFerriere on the vision systems, rapid deployment, and other system infrastructure. Ganesh Iyer contributed to the vision pipeline for Challenge 2, and Parv Parkhiya and Shubham Garg helped in the preliminary work toward Challenge 3. Azarakhsh Keipour and Rogerio Bonatti helped build the path generation and GPS-coordinate transform software, and provided valuable advice. We would also like to thank Lorenz Stangier for his preliminary work on the vision pipeline, Jay Maier and Valmiki Kothare for their quick and invaluable work on the ball-catching mount, and John Keller for helping get the systems up and running.

## ORCID

Anish Bhattacharya  https://orcid.org/0000-0002-5961-5486
Akshit Gandhi  https://orcid.org/0000-0001-8798-5321
Lukas Merkle  https://orcid.org/0000-0002-5432-1606
Rohan Tiwari  https://orcid.org/0000-0002-5412-2615
Karun Warrior  https://orcid.org/0000-0003-2488-005X
Stanley Winata  https://orcid.org/0000-0002-8519-3272
Andrew Saba  https://orcid.org/0000-0003-2462-2050
Kevin Zhang  https://orcid.org/0000-0003-3524-7719
Oliver Kroemer  https://orcid.org/0000-0003-2007-3867
Sebastian Scherer  https://orcid.org/0000-0002-8373-4688

# Appendices

## A. Rapid Deployment Details

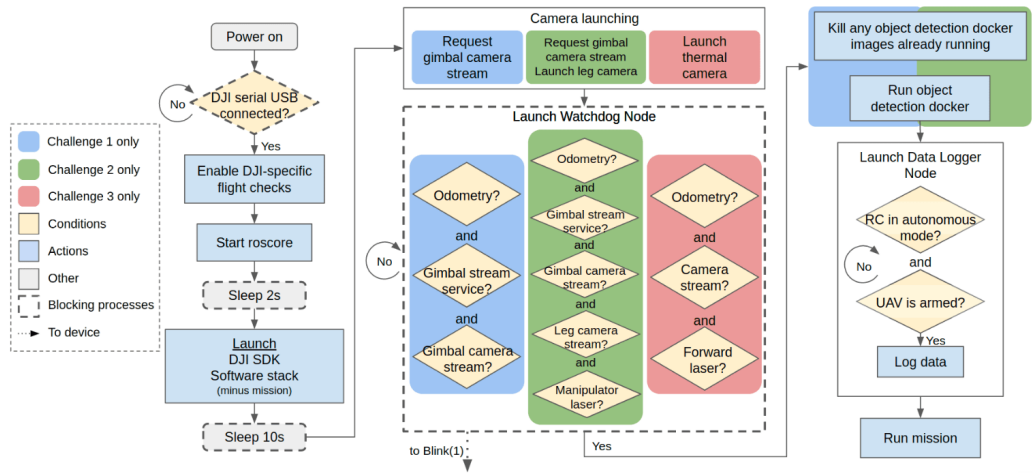

**Figure 24.** Flowchart presenting automated mission launch sequence across all UAV platforms. Challenge-specific components are color-coded according to the legend on the left.

The rapid-deployment pipeline (Figure 24) begins immediately upon system startup. First, the computer waits until the DJI serial device is registered. This is the connection through which DJI SDK data, such as odometry, system information, and commands are transferred to the UAV from the on-board computer. Afterwards, the system enables DJI-specific autonomous flight checks (e.g., allowing USB connection during flight, allowing power through the aircraft's rear external port), and starts roscore. A blocking two second sleep allows roscore to finish starting up, and then the DJI SDK ROS package is launched, as well as the core software stack seen in Figure 2 except for the mission state machine and external sensors. We enforce another pause in the system to allow these packages to start up. Next, we launch the sensors needed for the particular challenge. The gimbal cameras require the use of a DJI SDK rosservice call, the existence of which we use to determine if the DJI SDK has died. The Watchdog Node waits until the data streams of the odometry, gimbal camera, downward laser (encapsulated in the "Odometry" check), and additional laser sensors, are all active before continuing. This node also publishes success/failure of these streams to a Blink(1) USB LED, which was used as a visual indicator to determine the startup sequence status during remote, automated startup. The visual indication is very helpful to know the UAV mission status in-case there is no SSH connection made with the UAV on-board computer. Once these streams are detected, the pipeline continues to run a docker image for object detection in Challenges 1 and 2, and then runs a Data Logger Node which logs data if the remote control (RC) is switched to autonomous mode ("P mode" for the DJI M210) and the UAV has been armed. This is not a blocking operation; the mission node launches as soon as this node launches, and also includes a check for armed status to begin issuing takeoff commands to begin the run.

## B. Challenge 2: Adaptive Mask Segmentation

This section highlights our method to deal with the day-to-day changing visual conditions during testing as described in Section 9. We used a self-supervised approach to re-learn the placement structure segmentation logistic regression model during a single flight over the structure. This was done by using the density-based spatial clustering of applications with noise (DBSCAN) (Ester

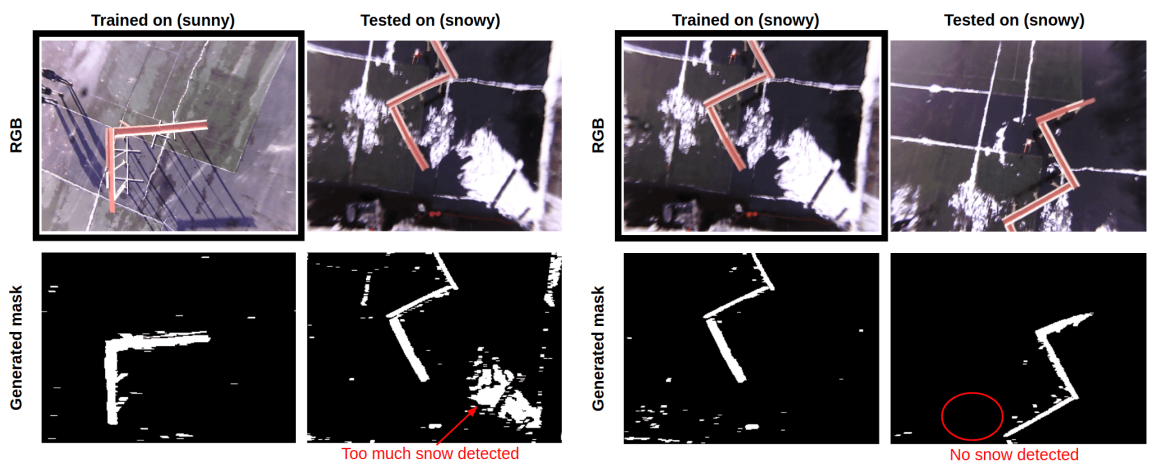

**(a)** Mask segmentation result when using a sunny day's logistic regression model.

**(b)** Mask segmentation result when automatically trained on a snowy day's image.

**Figure 25.** Adaptive placement structure mask segmentation when testing in variable conditions in Pittsburgh, PA for Challenge 2.

et al., 1996) clustering algorithm to create an initial segmented output, and these labels were then used to train the logistic regression model. Since DBSCAN is computationally expensive, we ran it once (requiring 5-10s to generate the labels). In order to select the cluster index that correspond to the structure, we took pixel values that we had collected from earlier runs and pushed them through the logistic regression model, selecting the most frequently occurring value. Figure 25 showcases the performance of the logistic regression model trained with different input images. We can see the model did not generalize well across images from different days, but this issue is solved by retraining the model each time. The main limitation of this approach is that we had to still tune the DBSCAN and logistic regression hyper-parameters at the beginning of each testing day, but this was much easier than manually labeling data.

While we developed this adaptive segmentation vision system that generalizes to the changing weather at the Pittsburgh test site, the competition arena scene remained very consistent across successive days, with almost no change in weather, lighting, or background. We therefore manually labeled images and trained a single logistic regression model which was used throughout the competition (we could have, equivalently, run the adaptive segmentation once). The outputs from this model were reliable and provided a good, consistent segmentation of the placement structure.

## C. Challenge 2: Visual Servoing Algorithm

Image-based visual servoing was used for block pickup and placement. The aim of visual servoing is to minimize an error $\mathbf{e}(t)$:

$$\mathbf{e}(t) = \mathbf{s}(t) - \mathbf{s}^* \tag{2}$$

where $\mathbf{s}(t)$ represents a vector of feature measurements from the image and $\mathbf{s}^*$ represents a vector of desired values. On deriving the relationship between the velocity of a 3-D point $\mathbf{X} = [X, Y, Z]$ in space and the corresponding velocity $\dot{\mathbf{x}}$ of its projection in the image $\mathbf{x} = [x, y]$, we arrive at Equation 3. A detailed derivation can be found in (Chaumette and Hutchinson, 2006). $\mathbf{v_c}$ represents a vector containing the linear and angular velocity of the camera which could be controlled using a velocity controller on the robot.

$$\dot{\mathbf{x}} = \mathbf{L_x} \mathbf{v}_c \tag{3}$$

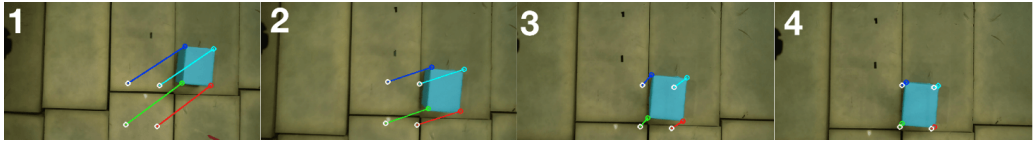

**(a)** Visual servoing over a block during a test in Pittsburgh, PA. Colored circles and lines represent the detected corners and pixel errors to the white circle goal coordinates.

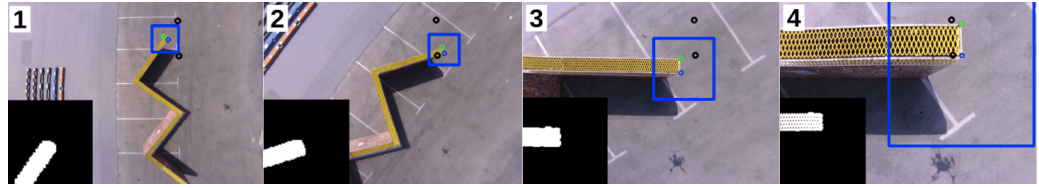

**(b)** Visual servoing over the placement structure in the MBZIRC arena. The mask of the updated tracker is shown in the bottom left. Black circles are the goal coordinates of the corners.

**Figure 26.** Image-based visual servoing progress for Challenge 2.

where $\mathbf{L_x}$ represents the interaction matrix corresponding to $\mathbf{x}$ and is expanded as

$$\mathbf{L_x} = \begin{bmatrix} -1/Z & 0 & x/Z & xy & -(1+x^2) & y \\ 0 & -1/Z & y/Z & 1+y^2 & -xy & -x \end{bmatrix} \tag{4}$$

Due to the underactuated model of the UAV, we only need to control linear velocities as well as yaw rate. This would mean we could reduce $\mathbf{v_c}$ to contain $[\dot{X}, \dot{Y}, \dot{Z}, \dot{\psi}]$ thus simplifying the interaction matrix to

$$\mathbf{L_x} = \begin{bmatrix} -1/Z & 0 & x/Z & y \\ 0 & -1/Z & y/Z & -x \end{bmatrix} \tag{5}$$

The goal is to use the relationship between the spatial velocity of the camera and the time varying error between the feature measurements from Equations 2 and 3 to develop a velocity control law.

$$\mathbf{v}_c = \lambda \mathbf{L_x}^+ \mathbf{e} \tag{6}$$

$\lambda$ represents a tunable gain parameter that can be used to appropriately scale the desired velocity of the camera. The desired $\mathbf{v_c}$ is then tracked using the UAV's velocity controller.

In our formulation, we chose to represent our features as pixel coordinates of detected feature points. Each point provides two constraints, $x$ and $y$ coordinates on the image coordinate frame, and hence a minimum of two points are required to fully constrain the system of equations to solve for the variables in the underactuated representation of $\mathbf{v_c}$. The individual interaction matrices corresponding to each measurement can be stacked and $\mathbf{v_c}$ can be solved using the pseudo-inverse of the combined interaction matrix. We detected the four corners of the white, ferromagnetic patch to align over the block, resulting in an over-constrained system, yielding a least squares estimate of $\mathbf{v_c}$. Servoing progress for pickup and placement can be seen in Figure 26.